\documentclass[11pt]{article}

\usepackage[final]{acl}

\usepackage{times}
\usepackage{latexsym}

\usepackage[T1]{fontenc}
\usepackage{booktabs}
\usepackage[utf8]{inputenc}

\usepackage{microtype}

\usepackage{inconsolata}

\usepackage{tabularx}
\usepackage{url}
\usepackage{graphicx}
\usepackage{multirow}
\usepackage{enumitem}
\usepackage{verbatim}
\usepackage{xcolor}
\usepackage[table]{xcolor}  % handles \cellcolor
\usepackage{colortbl}        % optional, xcolor table option covers most cases
\usepackage{tcolorbox}
\tcbuselibrary{breakable}

\usepackage[textsize=tiny]{todonotes}
\newcommand{\note}[4][]{\todo[author=#2,color=#3,size=\scriptsize,fancyline,caption={},#1]{#4}}

\newcommand{\yongxin}[2][]{\note[#1]{\textbf{Yongxin}}{orange!30}{#2}}
\newcommand{\Yongxin}[2][]{\yongxin[#1,inline]{#2}}

\usepackage{enumitem} % in preamble

\title{How Are Citations Wrong? A Taxonomy, Dataset, and LLM Analysis for Scientific Citation Misrepresentation}%we can rethink

\author{Qinyue Liu \quad Yongxin Zhou \quad François Portet \quad Cyril Labbé \\
  Univ. Grenoble Alpes, CNRS, Grenoble INP, LIG\\
  38000, Grenoble, France\\
  \texttt{firstname.lastname@univ-grenoble-alpes.fr}}
\begin{document}
\maketitle
\begin{abstract}
Citations are important in scientific communication, yet they are frequently misused, 
% ranging from subtle distortions of original research findings to AI-generated misinformation. 
either by distorting real sources or by fabricating non-existent ones.
% Existing work rarely focuses on fine-grained citation error taxonomies and explainable diagnosis. 
Existing work lacks both a fine-grained taxonomy of incorrect citations and resources for explanation annotation.
% To fill this gap, we introduce \textbf{VeriCite}, a manually annotated dataset of 1,453 citations, together with a taxonomy and benchmark for fine-grained citation classification. 
To fill this gap, we introduce \textbf{VeriCite}, comprising a taxonomy of incorrect citations, a manually annotated dataset of 1,453 citations, and a benchmark for fine-grained citation verification.
Our taxonomy defines nine distinct citation-error categories plus a \textit{Correct} label. Each incorrect citation is annotated with both an error category and a researcher-written explanation,
supporting future work on explainable detection. We benchmark eight LLMs spanning open-weight and closed-source paradigms, three parameter scales, and both thinking and no-thinking modes. Our analysis yields four findings: (i) open-weight and closed-source models perform comparably at the frontier, 
(ii) the benefit of thinking appears to be task-dependent, averaging
$+13.0$ macro-F1 on the ten-way taxonomy but only $+1.4$ when the ten-way taxonomy is reduced post hoc to a binary correct/incorrect decision, (iii) model scaling behavior is metric and mode dependent rather than universally beneficial, and (iv) error categories requiring semantic understanding remain the hardest across all models, and are also where thinking yields its largest gains. We release the taxonomy, the annotated dataset, the benchmark, and all evaluation scripts to support future work on citation verification and explainable citation verification.
\end{abstract}

\section{Introduction}
\begin{comment}

\Yongxin[]{Possible extensions: (1) identify the failure mode of models by comparing the reasoning traces against human-label explanations; (2) align the dataset to the real-world use case by identifying several relevant paragraphs in addition to the gold evidence paragraph, and report the  retrieval then generate pipeline results in appendix (sota retrieval model + sota llm or the best llm tested in the paper would be enough).}

\Yongxin[]{1. We need to rethink the title, LLM analysis is a bit too large. 2. better distinguish the difference with previous SCiCiteVal dataset, and should make use of the explanation annotations in the dataset as we listed it as something new and you stated as " explainable diagnosis". 3. check the feedback and weakness pointed by Stanford AI review and try to address them if possible. 4. check the usage of terms, make the same terms coherent throughout the paper, "citation error detection", "citation misrepresentation", "citation classification", "citation verification", "citation-error classification" seem to be the same thing in this paper. 5. add more references into Related Work. 6. read through the paper and check the story line, logical flow and transition phrases. 7. the figure could be more visible and consider change the second table results into one nice visualization while move the full table into appendix.}
\end{comment}

Citations are an important component of scientific publications, serving various functions that vary across disciplines \citep{citation_function2, citation_function}. Because citations carry these communicative functions, they have also become a measurable proxy for scientific impact. As bibliometric indicators have been widely adopted to evaluate research quality and influence \citep{metric_cite}, they have simultaneously fueled manipulation: researchers and organizations exploit citation counts to game evaluation systems \citep{game_metrics, Ioannidis2024}. This tension between the communicative role of citations and their evaluative role has brought the integrity of citation practices under increasing scrutiny.
\begin{figure}
  \centering
  \includegraphics[width=\columnwidth]{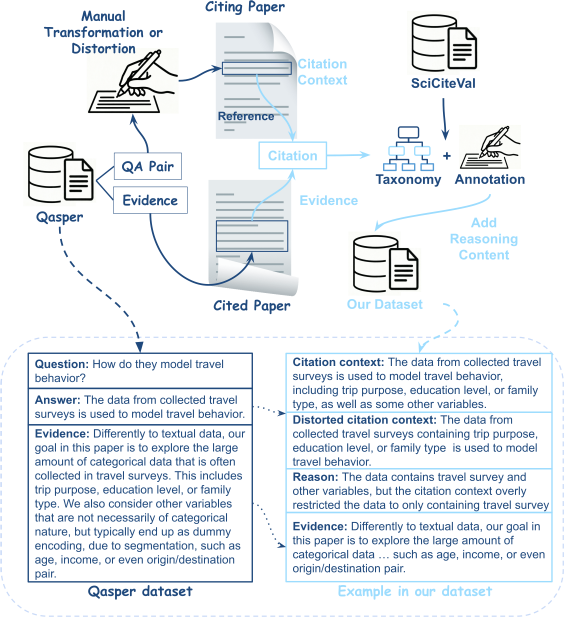}
    \caption{Dataset construction pipeline. 
    \textit{Bottom}: an \textit{Over-Specific} example where the distorted citation omits variables (age, income, origin/destination) present in the source evidence.}
  \label{fig:pipeline_example}
\end{figure}
Citation misconduct takes several forms. Paper mills represent one institutionalized form of this misconduct: organizations that sell authorship and citations for placement in legitimate journals \citep{Bishop2023}. Such citations are meaningless and deliberately detached from any connection to the citing work. A second and more recent form is \emph{AI-generated hallucination}, where the source does not exist at all: large language models fabricate plausible-looking references \citep{ravichander-etal-2025-halogen}, and such hallucinated citations have now been documented at scale in top NLP venues \citep{hallucination_acl}. Both forms can often be detected by verifying whether the referenced work exists or was genuinely produced. A subtler and more general issue is the case where a real, existing paper is cited, but the citation context distorts, overgeneralizes, or contradicts what the source actually reports. This form of error is harder to detect automatically and has received less systematic study.

Existing NLP work on citation verification has begun to address this challenge. Prior datasets typically formulate the task as fact-checking over a coarse label space: usually a binary or ternary supported/refuted/no-evidence scheme \citep{thorne-etal-2018-fever, wadden-etal-2020-fact}, and restrict evidence to abstracts \citep{wadden-etal-2020-fact, alvarez-etal-2024-zero} or structured elements such as tables and figures \citep{ho-etal-2025-table, wang-etal-2025-sciver}. These formulations share a common limitation: coarse labels do not characterize \textit{how} a citation misrepresents its source.

To address the gap, we introduce \textbf{VeriCite}, a manually annotated dataset for fine-grained textual citation verification built around a 9-category taxonomy of incorrect citation types, as well as a \textit{Correct} label for correct citations. Following this taxonomy, we construct the VeriCite dataset as illustrated in Figure~\ref{fig:pipeline_example}. A citation is treated as the combination of two elements: a \textit{citation context}: a complete sentence in the citing paper that contains at least one reference marker such as \texttt{[1]} or an author--year citation, and an \textit{evidence passage}: the content from the cited paper that should justify the claim expressed in the citation context. An incorrect citation is then defined as a citation for which the evidence passage does not support the claim made in the citation context. The importance of using citation context for verification is further motivated by \citet{context_cited}. Each incorrect citation is annotated with both an error category and a researcher-written explanation, enabling future work on explainable detection. After dataset construction, we evaluated several state-of-the-art LLMs on VeriCite and analyzed their reasoning traces. We established a systematic baseline for fine-grained citation
verification and exposed the limits of current models on this task.

Our main contributions are:
\begin{itemize}[noitemsep, topsep=2pt, leftmargin=*]
    \item A \textbf{taxonomy of incorrect citations}, covering different error types, from factual errors to errors requiring semantic understanding.
    \item A \textbf{manually annotated dataset} of 1,453 citations, validated by domain experts with cross-annotation on a held-out subset.
    \item \textbf{LLM baselines} with per-category performance analysis, revealing where current models systematically fail.
    \item \textbf{Explanation annotations} for incorrect citations, supporting future research on explainable verification and interpretable verification. These annotations can be used to better diagnose the LLM failure modes. %\yongxin[]{these annotations could be used to better diagnose the failure mode of llms}
\end{itemize}

\section{Related Work}

\subsection{Manual and Statistical Citation Verification}
Early work verified citations manually. \citet{kristof} analyzed citation errors from journal samples, \citet{old_verif} conducted a cross-disciplinary statistical survey of citation practices. More recently, \citet{jergas} systematically reviewed studies examining quotation errors in the medical literature. They defined quotation accuracy as whether a cited reference supports the statement of the citing author, which also aligns with our citation accuracy definitions. Their manual verification estimated a major quotation error rate of 11.9\%, a minor error rate of 11.5\%, and a total error rate of 25.4\%. Despite heterogeneity across studies, even the most conservative estimates remain substantial. For example, \citet{lawson1999} report 6.7\%, while \citet{davids2010} and \citet{goldberg1993} report total quotation error rates exceeding 80\%.

%These studies established the empirical scale of the problem: a meta-analysis reported a median quotation error rate of 20\% across accuracy studies \citep{wager2008}.

\subsection{Automated Citation and Claim Verification}
%With the development of large language models and scientific metadata systems, recent studies have leveraged these tools for citation and claim verification. 
We view citation verification and claim verification as two instantiations of the same underlying task: in both cases, a claim must be checked against a source document. Accordingly, NLP approaches to citation verification inherit their task formulation directly from scientific claim verification. These tasks treat the citation context as a claim and the cited work as its source, and ask whether the source supports the claim.

\textbf{Metadata-level citation verification.} \citet{sneaked} studied ``sneaked references'' used to manipulate citation counts by comparing metadata against reference lists. In a case study, they found at least 9\% of references were sneaked in while 56\% were lost from bibliometric platforms. A parallel line of work uses metadata matching and paper-database lookup to detect non-existing AI-hallucinated references.
This problem has reached top conferences in recent years, with nearly 300 papers at ACL, NAACL, and EMNLP in 2024--2025 found to contain at least one hallucinated citation \citep{hallucination_acl}, and with fabricated citations documented in 53 accepted NeurIPS 2025 papers\footnote{\href{https://fortune.com/2026/01/21/neurips-ai-conferences-research-papers-hallucinations/}{fortune.com/2026/01/21/neurips-ai-conferences-research-papers-hallucinations/}}. These works target \emph{invented} references, whereas we instead verify whether an existing reference is semantically consistent with its citing context.

\textbf{Text-based claim verification.}
\citet{wice} provides fine-grained entailment on sub-sentence units but targets Wikipedia claims rather than citation verification. \citet{distortion_twitter} annotates how findings shift when reported in news and tweets along four dimensions. Our work is narrower, operating at the level of a citation context against a specific evidence passage and offering more granular error types. Closest to ours, MissciPlus \citep{missciplus} pairs real-world misrepresented evidence with claims and classifies logical fallacies in health misinformation, and reports that LLMs struggle with such tasks.%\yongxin[]{are these all related to Scientific Literature (title of this subsec)?\textbf{Qinyue: These are not all from scientific literature, so I changed the title.}}

\textbf{Multimodal claim verification.}
%Beyond text, some studies \citep{sciclaimeval, wang-etal-2025-sciver} perform multimodal claim verification in scientific papers. They extract claims and evidence from existing papers to build datasets and then evaluate multimodal models on them. For example, \citet{sciclaimeval} introduced a cross-modal scientific claim verification dataset where refuted claims are created by perturbing evidence (figures and tables), rather than editing claims. \citet{wang-etal-2025-sciver} proposed a similar benchmark with 3,000 human-annotated examples covering four reasoning types, with refuted claims generated by perturbing entailed claims through semi-automated annotation.  Inspired by these work, we built our dataset using similar methods: the ``Refuted claims'' in their research correspond to our ``Distorted citations'' introduced in Section~\ref{dataset:construction}. However, our work concentrates on purely textual data rather than multimodal data.
Beyond text, a line of work extends claim verification to multimodal and multi-hop evidence \citep{sciclaimeval, wang-etal-2025-sciver, mmcv}. These studies build datasets by perturbing figures and tables into refuted claims \citep{sciclaimeval, wang-etal-2025-sciver}, or by generating multi-hop claims from question--answer pairs with LLM assistance and human feedback \citep{mmcv}. Our dataset shares their perturbation-based construction paradigm (Section~\ref{dataset:construction}): their ``refuted claims'' correspond to our ``distorted citations'' \citep{sciclaimeval, wang-etal-2025-sciver}, but we target fine-grained verification over purely textual evidence: models must assign each distorted citation to one of nine error types, which requires distinguishing semantically adjacent misrepresentations that a binary supported/refuted label cannot capture. Our dataset construction is also similar to \citet{mmcv} in transforming question--answer pairs into citation contexts (correspond to their ``claims''), but we produce all instances through manual annotation rather than LLM generation with human feedback, which we find necessary to ensure the reliability of our taxonomy.

%\yongxin[]{this paragraph should be more compact with more references, not detailed descriptions of two refs. also the last phrase states like we did similar while less than the previous work. \textbf{Qinyue: I just found another work that is good to be put here and included rag(similar to what we want to do next), paragraph modified, now it concentrates on comparison of work} }

\subsection{Citation Error Taxonomies}
The most closely related work in terms of taxonomy and methodology is \citet{Halil}, who construct a corpus of 3,063 citation instances in biomedical publications and combine sentence retrieval with fine-tuned claim verification models and LLM few-shot prompting. Their work is very meaningful to us. Their taxonomy has been built upon real incorrect citation cases in the biomedical domain. This makes their taxonomy match real-world incorrect citation scenarios. The real incorrect citation examples in their dataset also demonstrate the practical value of our taxonomy within human-written papers.

We adopt the categories and their definitions to refine our own taxonomy of incorrect citations. In addition to this work, our taxonomy is also refined by the taxonomy of a shared task on hallucination detection \citep{hallu_task}, with the explicit goal of covering error types arising in both human misrepresentation and LLM-assisted writing. Compared to prior taxonomies, our contribution offers two advances: (1) coverage of the machine learning domain rather than biomedical focus and (2) researcher-written explanations for each incorrect citation (an annotation layer absent from prior datasets) designed to support future work on explainable verification.

\section{Taxonomy and Dataset}

% ----------------------------------------------------------------------
\subsection{Taxonomy Construction}
\label{sec:taxonomy:construction}

\begin{comment}
    \paragraph{Motivation.}
Gathering naturally occuring \textit{Incorrect citations} is both
time-consuming and resource-intensive. To overcome this, we constructed a
controlled set of incorrect citations by systematically distorting correct
citation contexts. To do so in a principled and reproducible way, we needed a
structured set of distortion categories. We therefore developed a
\textit{draft taxonomy} of distortions that serves as a practical guide for
creating incorrect citations.\yongxin[]{I think the motivation of the taxonomy construction is not primarily for dataset construction, it's about better understand the real-world citation misrepresentation (as we stated), and we want to provide a dataset reflecting these real-world misusage and then we use our proposed taxonomy to construct a set of distorted citations.}

\yongxin[]{need one sentence here to state overall the three following phrases}

\paragraph{Draft taxonomy.}
Two researchers with NLP backgrounds independently distorted 20 citation
contexts using their corresponding evidence passages. They then compared
their distortions to identify common error patterns and defined an
initial set of categories.

\end{comment}

\paragraph{Motivation.}
Gathering naturally occurring incorrect citations is both time-consuming and resource-intensive. At the same time, the taxonomy we use to characterize incorrect citations should reflect the forms of their incorrectness that occur in practice, rather than being driven solely by the practical need to construct a dataset. We therefore develop a taxonomy of incorrect citations grounded in researcher judgment, existing taxonomies, and observations of naturally occurring incorrect citations. This taxonomy then serves as a principled guide for generating a controlled set of incorrect citations by systematically distorting correct ones (Section~\ref{dataset:construction}).

\paragraph{Draft taxonomy.}
Our first step is to build an initial draft taxonomy from a small set of exploratory distortions. Two NLP researchers with direct experience of academic writing and citation practices independently distorted 20 citation contexts using their corresponding evidence passages. They then compared their distortions to identify common error patterns and define an initial set of categories, which, together with the corresponding definitions, formed our draft taxonomy.

\paragraph{Theoretical refinement.}
To ensure conceptual rigor and broad coverage, we cross-referenced our taxonomy against established citation error frameworks from related work \citep{Halil, liu-etal-2026-sciciteval} and shared hallucination detection tasks \citep{hallu_task}. We observed overlaps between these frameworks and our draft taxonomy, so we kept the categories that were common and replaced certain categories with more explicit ones. For example, the ``Misinterpretation'' category in \citet{hallu_task} is general and overlaps with the ``Oversimplify'' category in \citet{Halil} as well as the ``Change Concept'' category in \citet{liu-etal-2026-sciciteval}. We therefore replaced it with two more explicit categories, \textit{Over-Generalization} and \textit{Over-Specific}, to represent the distinct semantic errors it conflated. These two categories separate the two distinct semantic errors that ``Misinterpretation'' conflated. For categories shared across frameworks, we refined or simplified our definitions accordingly.

\paragraph{Real-world validation.}
After finalizing the initial structure of our taxonomy, we actively searched published scientific papers for naturally occurring incorrect citations. We identified representative incorrect citations from the literature for every category except ``Other'', which is reserved for incorrect citations that do not fit into our taxonomy. We presented them in Appendix~\ref{tab:real-miscitations}, each accompanied by a comment explaining why it constitutes an incorrect citation.

%\Yongxin[]{should have one paragraph citing and explaining the final taxonomy Table \ref{tab:taxonomy}, explain a bit the different categories: require or not require semantic understanding, others cases hard to define, etc. (Is the example shown in "Other" category a correct citation?) \textbf{Done it, I think we had this before, but due to page limit, we had to delete it... Now as we only put this on arxiv, I think we are not very bothered by page limits.}}

\paragraph{Taxonomy overview.}
Our taxonomy classifies incorrect citations into nine categories according to \textit{how} and \textit{why} they misrepresent their source. The categories range from factual errors that are relatively easy to detect to semantic errors requiring deeper understanding of the cited evidence. Table~\ref{tab:taxonomy} provides full definitions and illustrative examples for each of the nine incorrect citation categories.

\begin{table*}[!ht]
\centering
\small
\renewcommand{\arraystretch}{1.5}
\begin{tabularx}{\textwidth}{@{}p{2.6cm} p{4.8cm} X@{}}
\toprule
\textbf{Category} & \textbf{Definition} & \textbf{Example} \\
\midrule
\multicolumn{3}{@{}l}{\textit{Does not require semantic understanding}} \\
\midrule
\textbf{Numeric Error}
  & Altering a key numerical value (percentage, quantity, measurement,
    year, etc.).
  & \textbf{Evidence:} 30\% of participants reported side effects. \newline
    \textbf{Citation Context:} The study found that 50\% of
    participants reported side effects. \\
\midrule
\textbf{Opposite}
  & Reversing the semantic polarity: positive~$\rightarrow$~negative, or
    negative~$\rightarrow$~positive.
  & \textbf{Evidence:} The treatment was found to have an impact on
    symptoms. \newline
    \textbf{Citation Context:} The treatment mentioned in this study
    had no effect on symptoms. \\
\midrule
\textbf{Tortured Phrase}
  & Replacing a standard scientific term with an odd, non-standard, or
    nonsensical phrase.
  & \textbf{Evidence:} The artificial intelligence technique that speaks
    to exploring a neural network has been created at an exceptional rate. \newline
    \textbf{Citation Context:} The human-made consciousness technique
    that speaks to exploring a neural organization has been created at
    an exceptional rate. \\
\midrule
\textbf{Entity Error}
  & Changing specific pre-defined details such as names, locations,
    institutions, as well as noun, verb, or adverb phrases, etc.
  & \textbf{Evidence:} We used Deepseek to verify the citations. \newline
    \textbf{Citation Context:} The authors used Chat-GPT to verify the
    citations. \\
\midrule
\multicolumn{3}{@{}l}{\textit{Requires semantic understanding}} \\
\midrule
\textbf{Over-Generalization}
  & The cited paper makes a specific finding. The citation context
    expands it to a broader, unjustified claim, and can be directly
    judged by the evidence.
  & \textbf{Evidence:} The sentiment analysis on Twitter faces
    difficulties like\ldots \newline
    \textbf{Citation Context:} The sentiment analysis on all the online
    social media faces difficulties like\ldots \\
\midrule
\textbf{Over-Specific}
  & Presenting a general finding as overly narrow or restricted, and can
    be directly judged by the evidence.
  & \textbf{Evidence:} The phrases from Twitter users' tweets and
    profile descriptions are extracted as the source of the textual
    data. \newline
    \textbf{Citation Context:} The authors extracted phrases containing
    emoticons from Twitter users' tweets and profile as their source of
    the textual data. \\
\midrule
\textbf{Partial Evidence}
  & The cited paper supports only part of the claim made in the citation
    context. The unsupported part cannot be judged directly by the
    evidence.
  & \textbf{Evidence:} The study finds that exercise improves sleep
    quality. \newline
    \textbf{Citation Context:} Exercise has been shown to improve sleep
    quality and reduce anxiety. \\
\midrule
\textbf{No Evidence}
  & The cited paper contains no statement that supports the specific
    claim being made.
  & \textbf{Evidence:} A paper discusses purely the history of AI
    without mentioning anything else. \newline
    \textbf{Citation Context:} According to the paper, the close-source
    models have better performance than the open-source models. \\
\midrule
\multicolumn{3}{@{}l}{\textit{Other situations}} \\
\midrule
\textbf{Other}
  & The citation doesn't belong to any mentioned category, or the error cannot be clearly classified into the given categories.
  & \textbf{Evidence:} We use three datasets for our experiments: IWSLT14 German-English, Turkish-English, and WMT14 English-German. \newline
    \textbf{Citation Context:} The authors used IWSLT14 German-English and IWSLT14 Turkish-English as the datasets for their experiments. \\
\bottomrule
\end{tabularx}
\caption{Our final taxonomy for the Incorrect citations, with definition of each category and an example. Categories are grouped into those that do not require semantic understanding, those that do, and other situations.}
\label{tab:taxonomy}
\end{table*}

\label{sec:dataset}

\subsection{Dataset Construction}
\label{dataset:construction}

\paragraph{Sources.}
We built our dataset from two existing sources: \textsc{SciCiteVal} \citep{liu-etal-2026-sciciteval} and \textsc{Qasper} \citep{qasper}.
\textsc{SciCiteVal} \citep{liu-etal-2026-sciciteval} provides correct and distorted citation contexts with aligned evidence passages, but formulates their classification as a three-way task: \emph{correct}, \emph{unrelated}, or \emph{incorrect}. The  \textsc{Qasper} dataset contains manually collected QA pairs with answers and evidence from NLP papers.

\paragraph{Role of the taxonomy.}
Our taxonomy serves two distinct roles depending on the source. For \textsc{SciCiteVal}, the taxonomy serves as the \emph{annotation standard}: annotators classify existing distorted citations into our categories. For \textsc{Qasper}, it serves as the \emph{generation guide}: annotators systematically distort correct citations according to the category definitions.

\paragraph{Processing \textsc{SciCiteVal} data.} 
For the \textsc{SciCiteVal} subset, two NLP researchers directly reviewed the provided \textit{distorted citation contexts} alongside their associated \textit{evidence passages}. Using our taxonomy as the standard, they manually annotated the specific error type for each instance and wrote a detailed explanation of why each distorted citation is incorrect. % The instances labeled ``Unrelated'' in \textsc{SciCiteVal} become ``No Evidence'' category in our dataset.

\paragraph{Processing \textsc{Qasper} data.}
For the \textsc{Qasper} subset, annotators executed a more complex three-step transformation, as illustrated in Figure~\ref{fig:pipeline_example}:
\begin{enumerate}
    \item \textbf{Reformulation.} Annotators read the original QA pairs and transformed them into \textbf{correct citation contexts}, ensuring that each citation accurately reflected its accompanying evidence passage. We treat each evidence passage as the cited content from the referenced paper that supports the transformed context.
    \item \textbf{Systematic distortion.} Following our taxonomy, annotators systematically distorted the correct citation contexts to generate \textbf{incorrect versions} that misrepresent the evidence. To maintain naturalness, they applied necessary fluency edits (e.g., adjusting word
    order, rephrasing) so that the distorted sentences remained
    grammatically coherent while logically diverging from the evidence.
    %Figure~\ref{fig:pipeline_example} shows an example of distorting a citation into the \textit{Over-Specific} category.
    \item \textbf{Explanation annotation.} For each distortion, annotators wrote a brief explanation of why the distorted citation is incorrect. 
    %This step supports future work on explainable citation verification.
\end{enumerate}

\paragraph{Scale and distribution.}
The resulting dataset contains 1,453 citations. The breakdown across our taxonomy is shown in Table~\ref{tab:nb_sample}.

\begin{table}
  \centering
  \small
  \begin{tabular}{lrr}
    \toprule
    \textbf{Category} & \textbf{N} & \textbf{Share (\%)} \\
    \midrule
    Correct             & 504 & 34.7 \\
    Entity Error        & 116 & 8.0 \\
    Numeric Error       & 85  & 5.9 \\
    Opposite            & 92  & 6.3 \\
    Tortured Phrase     & 47  & 3.2 \\
    Over-Generalization & 74  & 5.1 \\
    Over-Specific       & 61  & 4.2 \\
    Partial Evidence    & 25  & 1.7 \\
    No Evidence         & 431 & 29.7 \\
    Other               & 18  & 1.2 \\
    \midrule
    \textbf{Total}      & \textbf{1,453} & \textbf{100} \\
    \bottomrule
  \end{tabular}
  \caption{Distribution of citations across categories.}
  \label{tab:nb_sample}
\end{table}

\subsection{Inter-Annotator Validation}
As our dataset involves subjective judgments about citation correctness, we computed Inter-annotator agreement (IAA) separately for the distorted  citations of the two data sources.

\paragraph{SciCiteVal-derived portion (15 samples).}
  We randomly selected 15 citations covering all distortion categories.
  Two annotators independently labeled these 15 samples. Cohen's $\kappa$ was $0.763$.
\paragraph{Qasper-derived portion (49 samples).}
  Each annotator independently selected a set of samples that the other had distorted and re-annotated them without knowledge of the original labels (one annotator chose 24 samples, the other 25), yielding 49 samples that covered all distortion categories. The $\kappa$ was $0.634$.
\paragraph{Expert validation.}
A machine-learning expert then validated the annotations on the samples where the two annotators had agreed. The expert confirmed $83.3\%$ of the labels in the SciCiteVal-derived portion and $100\%$ of the labels in the Qasper-derived portion.%\yongxin[]{it could be nice to add two or three samples where the two annotators disagree (in the appendix or here), as annotator disagreement is also one thing important to note in this complex annotation task}

\paragraph{Interpretation.}
Both $\kappa$ values fall in the substantial range (0.61--0.80), indicating that our categories are operationalizable but retain some residual subjectivity inherent to fine-grained citation verification. Expert validation confirms that, where annotators agreed, labels are almost always correct. This indicates that the residual disagreement concentrates in genuinely borderline cases rather than systematic errors.

\section{Experimental Setup}
\label{sec:setup}

We evaluate eight LLMs on \textit{VeriCite}. The models are analyzed
along three dimensions: model paradigm (open-weight vs.\ closed-source), reasoning mode (thinking vs.\ no-thinking), and parameter scale within a single family. This section describes the models and evaluation protocol.

\begin{table*}[t]
\centering
\small
\setlength{\tabcolsep}{5pt}
\begin{tabular}{lcccc}
\toprule
& \multicolumn{2}{c}{\textbf{Multi-Class (macro / micro)}}
& \multicolumn{2}{c}{\textit{Binary (macro / micro, post-hoc)}} \\
\cmidrule(lr){2-3} \cmidrule(lr){4-5}
\textbf{Model}
  & \textbf{Thinking} & \textbf{No-Thinking}
  & \textit{Thinking} & \textit{No-Thinking} \\
\midrule
\multicolumn{5}{l}{\textit{Open-weight}} \\[2pt]
DeepSeek-V4-Flash
  & 51.94 / 75.48 & 34.25 / 69.67
  & \textit{95.26 / 96.91} & \textit{95.96 / 97.35} \\
DeepSeek-V4-Pro
  & 54.76 / 76.01 & 38.33 / 67.33
  & \textit{94.12 / 96.21} & \textit{91.42 / 94.67} \\
Qwen3-14B
  & 47.25 / 71.69 & 34.65 / 63.27
  & \textit{93.41 / 95.80} & \textit{89.76 / 93.75} \\
Qwen-3.6-35B
  & 49.37 / 76.60 & 37.31 / 71.67
  & \textit{\textbf{96.66} / \textbf{97.78}} & \textit{\textbf{96.93} / \textbf{97.96}} \\
Qwen-3.5 \textit{\small(397B)}
  & \textbf{60.08} / 79.53 & 42.45 / 68.85
  & \textit{95.38 / 96.99} & \textit{91.95 / 94.98} \\
\midrule
\multicolumn{5}{l}{\textit{Closed-source}} \\[2pt]
GPT-6.1-sol
  & 59.09 / 78.11 & {--} / {--}
  & \textit{93.19 / 95.66} & \textit{{--} / {--}} \\
Claude-Sonnet-5.5
  & 47.31 / 71.71 & \textbf{49.04} / \textbf{72.31}
  & \textit{91.80 / 94.86} & \textit{91.79 / 94.86} \\
Qwen-3.6-Plus
  & 58.48 / \textbf{79.97} & 42.17 / 72.25
  & \textit{96.43 / 97.64} & \textit{95.33 / 96.88} \\
\midrule
Average$^{\dagger}$
  & 53.54 / 76.14 & 39.74 / 69.34
  & \textit{94.53 / 96.48} & \textit{93.31 / 95.78} \\
\bottomrule
\end{tabular}
\caption{Macro-F1 / micro-F1 (\%) for ten-way taxonomy classification
(\textbf{Multi-Class}) and its post-hoc binary reduction
(\textit{Binary}, derived from the same predictions, as a coarse diagnostic). Best values per column in \textbf{bold}. Bootstrap confidence intervals are in Appendix~\ref{appendix:detailed results}. $^{\dagger}$No-Thinking averages exclude GPT-6.1-sol, which does not support a no-thinking mode. }
\label{tab:overall}
\end{table*}

\subsection{Models}
\label{sec:setup:models}

The eight models span four families and two deployment paradigms.

\paragraph{DeepSeek family.}
\textit{DeepSeek-V4-Flash} and \textit{DeepSeek-V4-Pro}~\citep{deepseek}
are open-weight models of different scales, both supporting thinking
and no-thinking modes.

\paragraph{Qwen family.}
We evaluate three Qwen models of increasing size: 
\textit{Qwen3-14B}~\citep{qwen3}, \textit{Qwen-3.6 35B}~\citep{qwen3.6-35}, and \textit{Qwen-3.5 397B}~\citep{qwen3.5}. All three support both reasoning modes.

\paragraph{Additional closed-source models.}
\textit{GPT-6.1-sol} (OpenAI)\footnote{\url{https://developers.openai.com/api/docs/models/gpt-6.1-sol}}
and \textit{Claude-Sonnet-5.5}
(Anthropic)\footnote{\url{https://platform.claude.com/docs/en/models/sonnet-5-5/overview}}
complete the evaluation set. GPT-6.1-sol does not expose a no-thinking
mode and is therefore evaluated in thinking mode only.

These eight models support three comparisons carried out in
Section~\ref{sec:results}: (i) open-weight vs.\ closed-source, using
the three representative open-weight models (DeepSeek-V4-Flash, DeepSeek-V4-Pro, Qwen-3.5 397B) against the three closed-source models (GPT-6.1-sol, Claude-Sonnet-5.5, Qwen-3.6-Plus); (ii) thinking vs.\ no-thinking, across the seven models that expose both modes; and (iii) scaling within the Qwen family, using Qwen3-14B, Qwen-3.6-35B, and Qwen-3.5 397B.

\subsection{Task and Prompting Protocol}
\label{sec:setup:task}

All models receive an identical prompt (Appendix \ref{app:prompt}) containing the \texttt{citation context},
the \texttt{evidence passage}, and the verbatim definition of each taxonomy
category, so that all models share the same label semantics. Models
return their answer in strict JSON format with two fields:
\texttt{predicted\_category} (one of the ten category names) and
\texttt{reason} (a short justification). Because the dataset is class-imbalanced, we report macro-F1 as the primary metric and micro-F1 for completeness.

\section{Results and Analysis}
\label{sec:results}

Table~\ref{tab:overall} summarizes the overall results. The ten-way multi-class task requires assigning one of nine incorrect categories or the \textit{Correct} label. We additionally report a binary view that collapses all nine error categories into a single \textit{Incorrect} class while keeping \textit{Correct} unchanged. This binary view is a post-hoc reduction of the same ten-way predictions rather than a separate task; it serves as a coarse-grained diagnostic that complements the fine-grained results.

The analysis is structured along four dimensions: open-weight vs.\ closed-source (Section~\ref{sec:analysis:open}), thinking vs.\ no-thinking (Section~\ref{sec:analysis:thinking}), scaling within the Qwen family (Section~\ref{sec:analysis:scaling}), and per-category attribution (Section~\ref{sec:analysis:percat}), followed by anomaly analysis (Section~\ref{anomaly}).

\subsection{Open-Weight vs.\ Closed-Source Models}
\label{sec:analysis:open}

We compare three representative open-weight models (DeepSeek-V4-Flash, DeepSeek-V4-Pro, Qwen-3.5 397B) against three closed-source models (GPT-6.1-sol, Claude-Sonnet-5.5, Qwen-3.6-Plus). The two metrics rank the top models differently. Under micro-F1, the best closed-source model (Qwen-3.6-Plus: 79.97) leads the best open-weight model (Qwen-3.5: 79.53) by 0.44 points. Under macro-F1, the ordering reverses, with Qwen-3.5 (60.08) leading GPT-6.1-sol (59.09) by 0.99 points. Among the closed-source models, Qwen-3.6-Plus and GPT-6.1-sol score 78--80 micro-F1, whereas Claude-Sonnet-5.5 scores 71.71. Open-weight models span 71.69--79.53 micro-F1, a range comparable to that of the closed-source models.

\begin{figure*}[t]
\centering
\includegraphics[width=\textwidth]{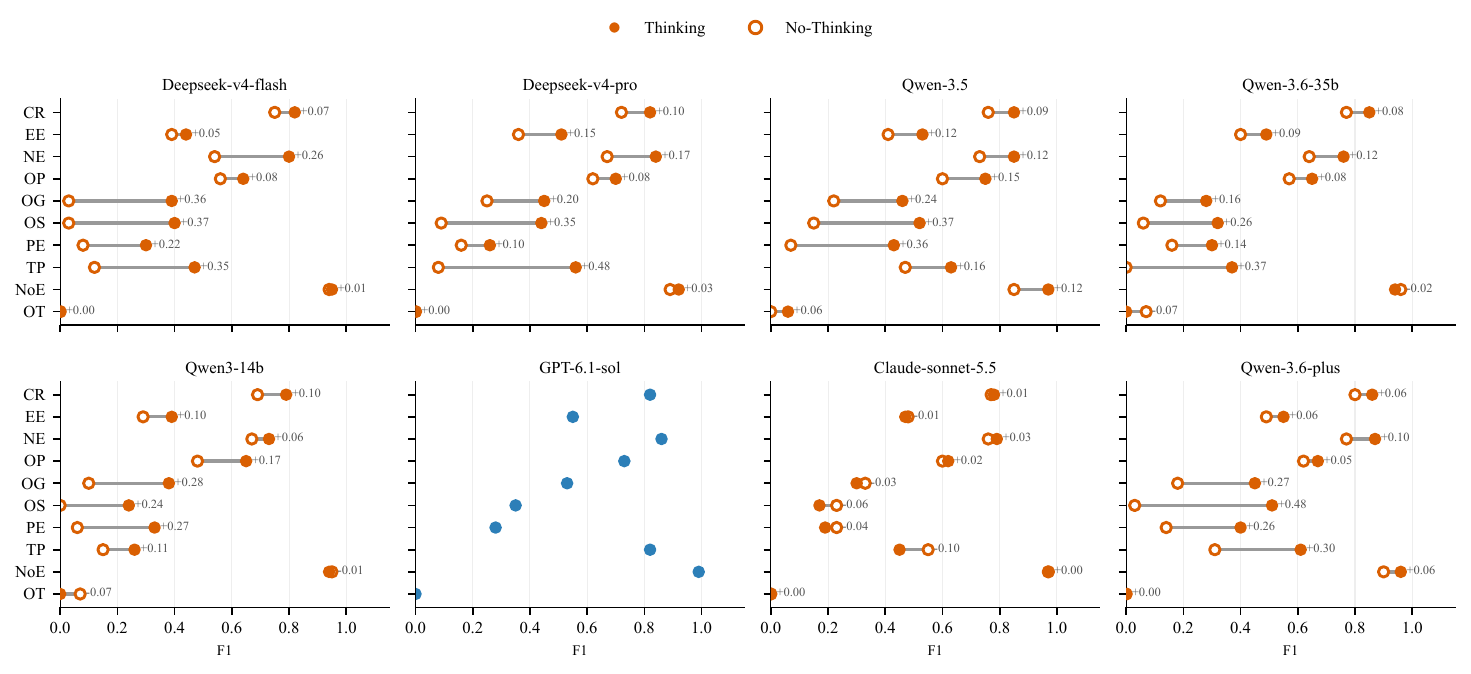}
\caption{Per-category Macro-F1 under thinking (filled) and no-thinking (hollow) modes, shown as dumbbells (one panel per model). Categories are ordered top-to-bottom as CR = Correct, EE = Entity Error, NE = Numeric Error, OP = Opposite, OG = Over-Generalization, OS = Over-Specific, PE = Partial Evidence, TP = Tortured Phrase, NoE = No Evidence, OT = Other. The horizontal position of each point encodes absolute F1 and the connecting line length encodes the thinking gain (annotated numerically to the right of each dumbbell). GPT-6.1-sol does not expose a no-thinking mode and is shown as filled points only.}
\label{fig:per-category-dumbbell}
\end{figure*}

%\yongxin[]{we need a more visible figure to present the results in Table \ref{tab:per-category} and move this table with detailed scores into appendix. \textbf{Qinyue: Just added a dumbbell and modified the method for analysis accordingly}}

\subsection{Thinking vs.\ No-Thinking}
\label{sec:analysis:thinking}

Across the seven models that expose both modes, thinking improves micro-F1 by $+6.5$ points on average and macro-F1 by $+13.0$ points on average. These gains are consistent across six of the seven models (micro-F1 range: $+4.93$ to $+10.68$; macro-F1 range: $+12.06$ to $+17.69$). Claude-Sonnet-5.5 is the exception and is analyzed separately in Section~\ref{anomaly}. 
The macro average gains roughly twice as much as the micro average, consistent with the gains concentrating in less frequent categories; Section~\ref{sec:analysis:percat} examines per-category F1.

Under the post-hoc binary reduction, all models exceed 89\% macro-F1 and 93\% micro-F1, and the mean thinking gain shrinks to $+1.4$ macro-F1 points (range $-0.70$ to $+3.65$), with two models turning negative (DeepSeek-V4-Flash: $-0.70$; Qwen-3.6-35B: $-0.27$). Since both views are derived from the same predictions, the contrast shows that thinking is most valuable when the label space requires distinguishing among closely related alternatives.

\subsection{Scaling within the Qwen Family}
\label{sec:analysis:scaling}

We compare three open-weight Qwen models of increasing size: Qwen3-14B, Qwen-3.6-35B, and Qwen-3.5 397B. Under thinking mode, both metrics increase monotonically: macro-F1 rises $47.25 \rightarrow 49.37 \rightarrow 60.08$ and micro-F1 rises $71.69 \rightarrow 76.60 \rightarrow 79.53$, a 12.8-point macro spread that confirms the ten-way taxonomy rewards model capacity. Under no-thinking mode, macro-F1 remains monotonic ($34.65 \rightarrow 37.31 \rightarrow 42.45$), but micro-F1 is non-monotonic: Qwen-3.6-35B (71.67) exceeds the larger Qwen-3.5 (68.85). 
This reversal appears only with thinking-mode off. Once thinking is enabled, the largest model leads on both metrics. Scaling therefore does not yield uniform gains, its effect depends on reasoning mode and on which metric is used.

\subsection{Per-Category Analysis}
\label{sec:analysis:percat}

Figure~\ref{fig:per-category-dumbbell} reports per-category F1 under thinking (filled) and no-thinking (hollow) modes, with each dumbbell's length representing the thinking gain. Detailed scores are reported in Appendix~\ref{appendix:detailed results}. We organize the discussion into two difficulty tiers and a separate set of categories where thinking helps most.

\paragraph{Easy tier (Average macro-F1 $>$ 0.75).} \textit{No Evidence} is the easiest category (thinking: 0.92--0.99), followed by \textit{Correct} (0.78--0.86) and \textit{Numeric Error} (0.73--0.87). All three show relatively high F1, and thinking yields little additional gain on them.

\paragraph{Hard tier (Average macro-F1 $<$ 0.5).} Five categories remain difficult across all models: \textit{Entity Error} (0.39--0.55), \textit{Over-Generalization} (0.28--0.53), \textit{Over-Specific} (0.17--0.52), \textit{Partial Evidence} (0.19--0.43), and \textit{Tortured Phrase} (0.26--0.82, with the widest variance). 
For \textit{Tortured Phrase} and \textit{Entity Error}, scores are lower than their apparent factual error level difficulty would suggest. Reasoning traces show that models tend to auto-correct the tortured phrase or entity error before judging the citation. Examples are in Appendix~\ref{appendix:case}.

\paragraph{Where thinking helps most.} Within the hard tier, the largest
thinking gains concentrate on \textit{Over-Generalization} and \textit{Over-Specific}: DeepSeek-V4-Flash improves by $+0.36$ and $+0.37$ respectively, and Qwen-3.5 by $+0.24$ and $+0.37$. Smaller but consistent gains appear on \textit{Opposite} and \textit{Partial Evidence}.

\subsection{Anomaly Analysis}
\label{anomaly}

The category \textit{Other} scores near 0.00 for almost all models, we attribute this to its low frequency and its comparatively ambiguous definition relative to the other categories.

Claude-Sonnet-5.5 is the only model whose thinking mode underperforms its no-thinking mode on the ten-way taxonomy. The deficit is modest: micro-F1 71.71 vs.\ 72.31, macro-F1 47.31 vs.\ 49.04. We reran the evaluation with an identical prompt and protocol; the rerun confirmed the result (macro-F1 = 47.71, within 0.40 of the original run; micro-F1 = 71.35, within 0.36), which argues against a sampling artifact. Comparing Claude-Sonnet-5.5 with the other frontier models category by category, its \textit{Correct} and \textit{Numeric Error} F1 (0.78 / 0.79) trail the strongest peers (Qwen-3.6-Plus: 0.86 / 0.87) by up to 8 points. 
The largest gaps are on \textit{Over-Specific} (0.17, vs.\ 0.24--0.52 for all other models), \textit{Partial Evidence} (0.19, vs.\ 0.26--0.43), and its \textit{Over-Generalization} score (0.30) is at the lower end of the range for other models (0.28--0.53). Its \textit{Entity Error} (0.47) and \textit{No Evidence} (0.97) scores are mid-range. The deficit thus falls mainly on categories that require fine-grained semantic discrimination. Concrete misclassified examples appear in Appendix~\ref{appendix:case}.

\section{Conclusion}

We introduce \textbf{VeriCite}, a taxonomy, dataset, and benchmark for fine-grained citation verification, with nine error categories plus a \textit{Correct} label. Benchmarking eight LLMs across two paradigms (open-weight vs.\ closed-source), three parameter scales, and both thinking and no-thinking modes yields several observations: parity between open-weight and closed-source frontier models, a thinking benefit that is task-dependent, and scaling behavior that depends on both metric and reasoning mode. Errors concentrate in categories requiring semantic understanding, where thinking also yields its largest gains. The VeriCite taxonomy and dataset are intended as open resources for citation verification and adjacent tasks such as textual entailment,
semantic similarity, and explainable NLI. We identify three directions for future work. First, we plan to investigate lightweight alternatives that can be deployed locally in place of large models, so that verification can still be served under privacy or economic constraints. Second, we aim to deepen the analysis of reasoning behavior by studying traces qualitatively and extending the taxonomy to new domains and languages. Third, we plan to collect more naturally occurring incorrect citations for broader comparison and analysis, so that our taxonomy better covers naturally occurring error types. Finally, because macro- and micro-F1 rank models differently on our data, we recommend that future citation-verification benchmarks report both metrics rather than a single number.

\section*{Limitations}
This work has several limitations. First, our dataset is moderate in size and incorrect examples were produced through controlled annotator distortion rather than collected from naturally occurring errors, so the error distribution may not fully reflect real-world patterns. However, each category is validated against at least one naturally occurring incorrect citations in published papers (Appendix~\ref{tab:real-miscitations}). Second, our dataset is predominantly drawn from the machine learning domain, so the
findings may not transfer directly to other scientific fields. Third, our benchmark covers only four model families (DeepSeek, Qwen, GPT, and Claude), limiting generalizability to other architectures and scales. Fourth, the post-hoc binary view has relatively high F1 (macro-F1 $>$ 89\%, micro-F1 $>$ 94\%), so it serves only as a diagnostic rather than as a discriminating benchmark. Finally, we evaluated models in a laboratory setting and have not yet tested their practical utility within real peer review or manuscript screening pipelines.%\yongxin[]{the fifth point is not a limitation of our work}

\bibliography{custom}

\appendix

%========================================================
% Appendix: real citation examples:
%========================================================
\section{Real Incorrect Citations for Each Category}

Table \ref{tab:real-miscitations} provides real, naturally occurring incorrect citations from the published literature for each category in our taxonomy, illustrating that the categories reflect errors observed in practice. For each example, we show the citation context and the corresponding evidence passage, followed by a comment explaining why it constitutes an incorrect citation. Four examples (Opposite, Numeric Error, Partial Evidence, and Entity Error) are drawn from the supplementary material of \citet{Halil}; the remaining examples were identified by us through manual search. The \textit{Other} category is omitted, as it is reserved for errors that do not fit the taxonomy, and the \textit{Tortured Phrase} example is presented as a phrase rather than as a single citation--evidence pair, since tortured phrases are defined at the phrase level.

\begin{table*}[ht]
\centering
\footnotesize
\begin{tabularx}{\textwidth}{@{}p{2.2cm}X@{}}
\toprule
\textbf{Category} & \textbf{Example} \\
\midrule
\textbf{Over-Specific} &
\textbf{Citation Context:} We conducted three focus groups to collect data, a number recommended in recently published studies [25, 26]. \newline
\textbf{Evidence Passage:} We identified 220 papers published in 117 journals. In these papers insufficient reporting of sample sizes was common. The number of focus groups conducted varied greatly (mean 8.4, median 5, range 1 to 96). Studies with high numbers of focus groups did not offer explanations for number of groups. \newline
\textbf{Comment:} The authors did not recommend the number of focus groups; they merely performed statistical research. The authors of the cited paper later stated: ``In no part of our study did we recommend specific sample sizes.'' They also noted: ``We assessed 205 articles that had referred to our 2011 study. We identified the type of quotation error we were interested in, namely the misrepresentation of descriptive information as normative, in 50.7\% of the included articles.'' \\
\midrule

\textbf{Over-Generalization} &
\textbf{Citation Context:} New developments in science [69--73, \textbf{74}, 75] and technology [76] have significant impact on human health [77--81] and life [82, 83]. \newline
\textbf{Evidence Passage:} Cement manufacture is one of the reasons for society's increasing carbon dioxide footprint. The development of a sustainable construction material is therefore needed to replace Portland cement fully or partially in building construction applications. Geopolymer concrete is a sustainable cementitious material, which is claimed to reduce carbon dioxide emissions and utilise waste materials such as fly ash, metakaolin and blast-furnace slag. Fly-ash-based geopolymer concrete with an activating solution of a mixture of silica fume, sodium hydroxide and water was investigated. Four Portland cement replacement weight ratios (0\%, 5\%, 10\% and 15\% by weight of fly ash) were studied. The effects of the Portland cement replacement on the early geopolymerisation process, compressive strength, modulus of elasticity and Poisson's ratio were investigated. Acoustic emission monitoring results showed that the early geopolymerisation process was enhanced when Portland cement replacement was increased. The compressive strength and modulus of elasticity were significantly increased when the Portland cement ratio increased, while Poisson's ratio was adversely affected. \newline
\textbf{Comment:} The citation applies only to the preceding terms, which are so broad in scope that, technically speaking, almost any paper dealing with technological development could be cited. \\
\midrule

\textbf{No Evidence} &
\textbf{Citation Context:} The contribution of organic materials has been well acknowledged in the application of electronic devices [28--32]. \newline
\textbf{Evidence Passage:} Male moths compete to arrive first at a female releasing pheromone. A new study reveals that additional pheromone cues released only by younger females may prompt males to avoid them in favor of older but more fecund females. \newline
\textbf{Comment:} The authors of the cited paper later published an article discussing these completely off-topic citations that reference their earlier work. An error in the DOI resolution mechanism caused the paper to be cited by thousands of unrelated studies. \\
\midrule

\textbf{Opposite} &
\textbf{Citation Context:} This is coherent with the fact that hACE2 expression were not observed in the gut of the mice used in that study [37]. \newline
\textbf{Evidence Passage:} In the gastrointestinal tract of K18-hACE2 mice, hACE2 was expressed most abundantly in the colon, which correlated with infection seen at later time points. \newline
\textbf{Comment:} The example was found in the supplementary material of \citet{Halil}. \\
\midrule

\textbf{Numeric Error} &
\textbf{Citation Context:} Recent reviews show 32\% loss to follow-up during this pre-treatment period despite eligibility for ART [12]. \newline
\textbf{Evidence Passage:} The rate of early mortality and loss to follow-up, which itself portends mortality for many, averages 23\% across the region [2]. \newline
\textbf{Comment:} The example was found in the supplementary material of \citet{Halil}. \\
\midrule

\textbf{Partial Evidence} &
\textbf{Citation Context:} Most studies have focused on the prevalence of high BMI, associating it with increased risk of conditions such as coronary heart disease, stroke, diabetes, and cancer. \newline
\textbf{Evidence Passage:} Metabolic mediators of the effects of body-mass index, overweight, and obesity on coronary heart disease and stroke: a pooled analysis of 97 prospective cohorts with 1\(\cdot\)8 million participants. \newline
\textbf{Comment:} The example was found in the supplementary material of \citet{Halil}. The authors explained that ``BMI-cancer association is not examined in the referenced paper.'' \\
\midrule

\textbf{Tortured Phrase} &
\textbf{Example:} The belief of the have a look at was that \textit{device getting to know algorithms} showed terrific capability in offering correct and timely information analysis while optimizing real-time programs. \newline
\textbf{Comment:} The \textit{Machine learning algorithms} is replaced by the tortured phrase: \textit{device getting to know algorithms}.\\ % source: https://pubpeer.com/publications/D36053FAA3E8D5686ABFBE88BDE25C#0
\midrule

\textbf{Entity Error} &
\textbf{Citation Context:} Diabetic Macular Oedema (DMO) is one of the leading causes of blindness in the working-age population [1, 2]. \newline
\textbf{Evidence Passage:} Diabetic retinopathy (DR) is the leading cause of blindness among working-aged adults around the world (1). \newline
\textbf{Comment:} The example was found in the supplementary material of \citet{Halil}. The authors explained that the evidence passage mentioned Diabetic Retinopathy (DR), but the citation context misused the entity as Diabetic Macular Oedema (DMO). \\
\bottomrule
\end{tabularx}
\caption{
% Real incorrect citations in the scientific literature that fit into our taxonomy.
One real incorrect citation from the published literature for each category in our taxonomy, with the citation context, evidence passage, and a comment explaining the error. The \textit{Other} category is omitted because it is reserved for errors outside the taxonomy.
}
\label{tab:real-miscitations}
\end{table*}

\clearpage
% =========================================================
% Appendix: Annotation Guideline
% =========================================================
\section{Annotation Guideline}
\label{app:annotation}

The following is the annotation guideline provided verbatim to the annotators for the Qasper-derived portion of VeriCite. In the guideline, \textit{twisted} is used interchangeably with \textit{distorted}. The guideline is shown in Table~\ref{tab:annotation-guideline}.
The distortion taxonomy referenced at the end of the guideline is reproduced in Table~\ref{tab:taxonomy}.

\begin{table*}[!ht]
\centering
% \small
\begin{tcolorbox}[
    title={\textbf{Annotation Guideline}},
    breakable,
    colback=gray!5,
    % colframe=cyan!35,
    colframe=cyan!35!gray!40,
    boxrule=0.4pt,
    left=6pt, right=6pt, top=4pt, bottom=4pt,
    fonttitle=\bfseries,
]
\small

\noindent\textbf{Background}\\
You are going to annotate the Qasper-derived part of the VeriCite
dataset, which is extracted from the Qasper dataset, a human-annotated
dataset including questions and answers, as well as evidence within
NLP-field scientific papers. This guideline explains the structure of
the annotation file and gives instructions for the annotator, followed
by a detailed distortion taxonomy including the category name,
definition, and example. Please read the guideline carefully before
annotation.

\medskip
\noindent\textbf{The Annotation File}\\
The columns in the file are as follows.

\smallskip
\noindent\textit{Columns that should not be modified.}
\begin{itemize}[leftmargin=*, itemsep=2pt, topsep=2pt]
  \item \textbf{id}: The id of the paper in the Qasper dataset.
  \item \textbf{title}: The title of the paper in the Qasper dataset.
  \item \textbf{question}: The question proposed by the Qasper annotator
        after reading the paper.
  \item \textbf{answer}: The answer to the question proposed.
  \item \textbf{evidence}: The evidence that can justify the answer to
        the question.
\end{itemize}

\smallskip
\noindent\textit{Columns that you need to contribute.}
\begin{itemize}[leftmargin=*, itemsep=2pt, topsep=2pt]
  \item \textbf{Citation\_context}: The citation context transformed from
        the Qasper dataset by the annotator.
  \item \textbf{Twisted\_context}: The incorrect citation context distorted
        by the annotator according to our taxonomy.
  \item \textbf{Category}: The category of the twisted citation context.
  \item \textbf{Twist\_reason}: The reason that explains why the twisted
        citation context is incorrect.
  \item \textbf{Comment}: The annotator's comment (optional).
\end{itemize}

\medskip
\noindent\textbf{Task Description}\\
In the annotation file, for each sample, you should first read and
understand:
\begin{itemize}[leftmargin=*, itemsep=2pt, topsep=2pt]
  \item[A.] The question and its corresponding answer.
  \item[B.] The evidence that justifies the answer.
\end{itemize}
After reading, your task is to:
\begin{enumerate}[leftmargin=*, itemsep=2pt, topsep=2pt]
  \item Verify if A can be transformed into a citation context that
        correctly cites B. If yes, transform A into a citation context,
        with minimal modifications if needed for fluency, and write it
        down in the \textit{Citation\_context} column. If no, exclude
        this sample.
  \item Distort the transformed citation context to make it an incorrect citation according to our taxonomy defined below. Then enter the distorted content in the \textit{Twisted\_context} column and the distortion category in the \textit{Category} column.
  \item Take notes of how you distorted the citation context and why it
        becomes an incorrect citation by marking the changed part in red, and
        write down the reason in the \textit{Twist\_reason} column.
\end{enumerate}

Before you begin distorting the citation context, please read the distortion taxonomy carefully.

The distortion taxonomy is given in Table~\ref{tab:taxonomy}.

\end{tcolorbox}
\caption{Annotation guideline provided verbatim to the annotators for the Qasper-derived portion of VeriCite.}
\label{tab:annotation-guideline}
\end{table*}

%========================================================
% Appendix: Real citation analysis:
%========================================================
\section{Mis-classification Case Study}
\label{appendix:case}

\subsection{Tortured Phrase Case Study}
\begin{table*}[ht]
\centering
\footnotesize
\setlength{\tabcolsep}{5pt}
\renewcommand{\arraystretch}{1.1}
\begin{tabular}{@{}p{0.13\textwidth} p{0.82\textwidth}@{}}
\toprule
\textbf{Category} & \textit{Tortured Phrase} \\
\midrule
\textbf{Evidence}
& Some scalp infections may be treatable if diagnosed early. Some but not all diseases may go on their own. Only an expert physician can detect the illness by visual observation. \ldots\ Although early-stage detection of hair and scalp-related diseases is the key to the treatment process, hair loss and scalp diseases can often go undetected due to a lack of awareness and a lengthy diagnosis test. \\
\midrule
\textbf{Citation Context}
& An AI-based application might pave the way to facilitate early disease detection. Disease detection using \textbf{device learning} approaches is gaining popularity in health informatics. Therefore, AI-based approach for fast detection is a good way to make people go for the early diagnosis of the hair diseases. \\
\midrule
\textbf{Annotator Explanation}
& The tortured phrase \textbf{``device learning''} appears in the citation context, where the correct form should be ``machine learning''. \\
\midrule
\textbf{Qwen-3.6-Plus Reasoning}
& ``The citation context is fully justified by the evidence. The evidence contains verbatim matches for the first two sentences and explicitly supports the remaining claims regarding the rising popularity of machine learning (referred to as `device learning' in the context, \emph{a minor typo}), as well as the capability of AI-based applications to enable fast, early diagnosis of hair and scalp diseases.'' \\
\midrule
\textbf{Prediction}
& \textit{Correct} \\
\bottomrule
\end{tabular}
\caption{A \textit{Tortured Phrase} case misclassified by Qwen-3.6-Plus as \textit{Correct}. The model detects the distorted phrase ``device learning'' but dismisses it as ``a minor typo'', overlooking that the distortion
\emph{is} the error.}
\label{tab:case-tp}
\end{table*}

\begin{table*}[ht]
\centering
\footnotesize
\setlength{\tabcolsep}{5pt}
\renewcommand{\arraystretch}{1.1}
\begin{tabular}{@{}p{0.13\textwidth} p{0.82\textwidth}@{}}
\toprule
\textbf{Category} & \textit{Numeric Error} \\
\midrule
\textbf{Evidence}
& ``Compared with 2015, our QA-SRL gold yielded 1593 annotations, with 989 core and 604 adjuncts, while theirs yielded 1315 annotations, 979 core and 336 adjuncts.'' \\
\midrule
\textbf{Citation Context}
& ``Compared to the previous dataset, the new dataset coverages 300 more annotations.'' \\
\midrule
\textbf{Annotator Explanation}
& The new dataset actually covers \textbf{278} more annotations, not 300. \\
\midrule
\textbf{Claude-Sonnet-5.5 Reasoning}
& ``The evidence reports 1593 annotations in the new gold set against 1315 in the previous one, a difference of \textbf{278}, which is about 300 more annotations. The citation's figure is \emph{a reasonable approximation}.'' \\
\midrule
\textbf{Prediction}
& \textit{Correct} \\
\bottomrule
\end{tabular}
\caption{A \textit{Numeric Error} case misclassified by Claude-Sonnet-5.5 as \textit{Correct}. The model computes the correct difference (278) but explicitly accepts ``300'' as ``a reasonable approximation'', thereby absolving the citation of its numerical imprecision.}
\label{tab:case-ne}
\end{table*}

\begin{table*}[ht]
\centering
\footnotesize
\setlength{\tabcolsep}{5pt}
\renewcommand{\arraystretch}{1.1}
\begin{tabular}{@{}p{0.13\textwidth} p{0.82\textwidth}@{}}
\toprule
\textbf{Category} & \textit{Correct} \\
\midrule
\textbf{Evidence}
& ``Given that there is no public dataset available with financial intents in Portuguese, we have employed the incremental approach to create our own training set for the Intent Classifier. We have created domain-specific word vectors by considering a set 246,945 documents, corresponding to of 184,001 Twitter posts and 62,949 news articles, all related to finance.'' \\
\midrule
\textbf{Citation Context}
& ``A custom dataset with user questions, a set of documents, Twitter
posts, and news articles, all related to finance, is used in the study.'' \\
\midrule
\textbf{Annotator Explanation}
& The evidence confirms that the authors created their own training set (custom dataset) for an Intent Classifier, which implies user
questions/intents. Because the evidence is only implicit about ``user
questions'', the annotators flagged this case as \emph{borderline}. \\
\midrule
\textbf{GPT-6.1-sol Reasoning}
& ``The evidence supports creating a custom training set and using
finance-related documents comprising Twitter posts and news articles.
However, it does not state that the dataset includes \textbf{user
questions}.'' \\
\midrule
\textbf{Prediction}
& \textit{Partial Evidence} \\
\bottomrule
\end{tabular}
\caption{A \emph{borderline} case labeled \textit{Correct} by annotators but predicted as \textit{Partial Evidence} by GPT-6.1-sol. Annotators reasoned that ``Intent Classifier'' implicitly entails user questions, whereas the model treated the missing explicit mention as a gap. The dataset splits 3--5 across the eight evaluated models, making this one of the most contested instances in VeriCite.}
\label{tab:case-fp}
\end{table*}

Table~\ref{tab:case-tp} presents a representative failure on the
\textit{Tortured Phrase} category. The citation context paraphrases the evidence passage and, on the surface, appears well supported: the claim that AI-based approaches facilitate early disease detection is consistent with the evidence, and the model correctly identifies this overlap.
However, the context substitutes \textit{``device learning''} for
\textit{``machine learning''}, a phrase-level distortion that is precisely what the \textit{Tortured Phrase} category is designed to capture.

The model does detect the anomaly: its reasoning explicitly notes that ``machine learning'' is ``referred to as `device learning' in the context''. However, the model explicitly dismisses the distortion as ``a minor typo'' and assigns the \textit{Correct} label. This suggests that models tend to treat lexical anomalies as surface noise rather than as semantically meaningful distortions, an assumption that is reasonable for spelling errors but incorrect for the \textit{Tortured Phrase} category, where the substitution is the error.

\subsection{Numeric Error Case Study}
\label{sec:appendix:case-ne}

Table~\ref{tab:case-ne} shows a complementary failure mode on the
\textit{Numeric Error} category. Here the model performs the arithmetic correctly, it explicitly computes the difference between 1593 and 1315 as 278, yet still assigns the \textit{Correct} label, on the grounds that the citation's figure of 300 is ``a reasonable approximation''.

Two aspects of this case are notable. First, the failure is not
arithmetic but normative: the model knows the right number and
still accepts the wrong one. This distinguishes \textit{Numeric Error} from a pure calculation failure and shows that the category demands a strict standard of numerical fidelity that models do not spontaneously adopt. Second, the model's justification (``about 300'') reveals an approximation bias: when the citation's number is close to the correct value, the model relaxes its criterion and treats the discrepancy as acceptable rounding rather than as an error. This example suggests that certain models tend to normalize away small surface-level deviations rather than treating them as the substantive error that the taxonomy is designed to capture.

\subsection{Borderline Correct Case Study}
\label{sec:appendix:case-fp}

Table~\ref{tab:case-fp} presents a case labeled \textit{Correct} by
annotators but predicted as \textit{Partial Evidence} by GPT-6.1-sol. The evidence describes a custom training set for an \emph{Intent Classifier}, and the citation summarizes it as ``a custom dataset with user questions, a set of documents, Twitter posts, and news articles.'' Annotators reasoned that an Intent Classifier necessarily operates on user intents or questions, so the citation's mention of ``user questions'' is entailed by the evidence even without a verbatim match. The model instead treated the missing explicit mention as an evidential gap and downgraded the citation. In this case, the model applies a verbatim-match standard, while annotators apply a semantic-entailment standard. Notably, the eight models are nearly split on this instance, with three predicting \textit{Correct} and five predicting an error category. This split suggests that future versions of the taxonomy would benefit from an explicit rule for implicit entailment.

% =============================================================
% Prompt Sample
% =============================================================

\section{Prompt Sample}

Table~\ref{app:prompt} shows the full prompt template used across all models and conditions (Section~\ref{sec:setup}). Category definitions are provided verbatim so that all models share identical label semantics.

\begin{table*}[ht]
\centering
\begin{minipage}{\textwidth}

% \vspace{0.8em}
% \textbf{Prompt Sample:}

% We use a single prompt template across all models and all experimental
% conditions (Section~\ref{sec:setup}). The template is shown below.
\begin{tcolorbox}[
  title={Prompt Template\label{prompt:zero-shot}},  % Put label inside title
  fonttitle=\bfseries,
    colback=gray!5,
    % colframe=cyan!35,
    colframe=cyan!35!gray!40,
    boxrule=0.4pt,
    arc=2pt,
    left=8pt, right=8pt, top=8pt, bottom=8pt,
    %breakable,
]
\small

\noindent\textbf{Instruction:}

The user will provide you a `citation context' and an `evidence'. The
`citation context' actually cites the `evidence'. Please analyze them, and
verify whether the content described in the `citation context' can be
justified by the `evidence' or not. If the `evidence' justifies the
`citation context', the citation is correct; otherwise, it is incorrect.

\vspace{0.6em}
\noindent\textbf{The definition of each incorrect category:}
\begin{itemize}[leftmargin=1.4em, itemsep=1pt, topsep=3pt, parsep=0pt]
    \item \textbf{Numeric Error:} Altering a key numerical value
    (percentage, quantity, measurement, year etc.).
    \item \textbf{Opposite:} Reversing the semantic polarity: positive
    $\rightarrow$ negative, or negative $\rightarrow$ positive.
    \item \textbf{Tortured Phrase:} Replacing a standard scientific term
    with an odd, non-standard, or nonsensical phrase.
    \item \textbf{Entity Error:} Changing specific pre-defined details such
    as names, locations, institutions, nouns, verbs, adverbs etc.
    \item \textbf{Over-Generalization:} The cited paper makes a specific
    finding. The citation context expands it to a broader, unjustified
    claim, and can be directly judged by the evidence.
    \item \textbf{Over-Specific:} Presenting a general finding as overly
    narrow or restricted, and can be directly judged by the evidence.
    \item \textbf{Partial Evidence:} The cited paper supports only part of
    the claim made in the citation context. The unsupported part cannot be
    judged directly by the evidence.
    \item \textbf{No Evidence:} The cited paper contains no statement that
    supports the specific claim being made.
    \item \textbf{Other:} The citation doesn't belong to any mentioned
    category, or the error cannot be clearly classified into the given
    categories.
\end{itemize}

\vspace{0.6em}
\noindent\textbf{Output format.}
Only output valid JSON format. Do not output anything else, and do not
rename keys.

\begin{quote}
\ttfamily\small
\{\\
\hspace*{2em}"Prediction": "Output the category of the citation here",\\
\hspace*{2em}"Reason": "Output the reason here"\\
\}
\end{quote}

\vspace{0.6em}
\noindent\textbf{Input.}\\
\textbf{Citation context:} \texttt{\{Citation context\}}\\
\textbf{Evidence:} \texttt{\{Evidence passage\}}
\end{tcolorbox}
\end{minipage}
% \caption{Prompt template used across all models and conditions.}
\caption{Prompt template used across all models and all experimental
conditions. Category definitions are provided verbatim so that all models
share identical label semantics.}
\label{app:prompt}
\end{table*}

\newpage
\section{Detailed Results}
\label{appendix:detailed results}
Table~\ref{tab:ci} reports 95\% bootstrap confidence intervals
(1,000 resamples) for the macro-F1 scores in Table~\ref{tab:overall}. Table~\ref{tab:per-category} reports detailed per-category macro-F1 and micro-F1 scores under thinking and no-thinking modes of all models.

\begin{table*}[t]
\centering
\small
\setlength{\tabcolsep}{6pt}
\begin{tabular}{lcccc}
\toprule
& \multicolumn{2}{c}{\textbf{Multi-Class}} & \multicolumn{2}{c}{\textbf{Binary}} \\
\cmidrule(lr){2-3} \cmidrule(lr){4-5}
\textbf{Model} & \textbf{Thinking} & \textbf{No-Thinking}
& \textbf{Thinking} & \textbf{No-Thinking} \\
\midrule
Deepseek-v4-flash   & 51.94 [48.81, 55.05] & 34.25 [31.94, 36.74]
                & 95.26 [94.24, 96.22] & 95.96 [95.01, 96.83] \\
Deepseek-v4-pro & 54.76 [51.79, 57.78] & 38.33 [35.90, 40.97]
                & 94.12 [92.93, 95.14] & 91.42 [90.02, 92.70] \\
GPT-6.1-sol     & 59.09 [56.17, 61.89] & -- / --
                & 93.19 [91.90, 94.33] & -- / -- \\
Claude-Sonnet-5.5 & 47.31 [44.35, 50.24] & 49.04 [46.03, 51.89]
                & 91.80 [90.41, 93.07] & 91.79 [90.38, 93.09] \\
Qwen-3.6-Plus        & 58.48 [55.14, 61.54] & 42.17 [39.36, 44.92]
                & 96.43 [95.58, 97.26] & 95.33 [94.36, 96.24] \\
Qwen-3.5        & 60.08 [56.80, 63.47] & 42.45 [39.73, 45.26]
                & 95.38 [94.43, 96.30] & 91.95 [90.63, 93.19] \\
Qwen-3.6-35b    & 49.37 [45.82, 52.62] & 37.31 [34.39, 40.43]
                & 96.66 [95.84, 97.40] & 96.93 [96.11, 97.63] \\
Qwen3-14b       & 47.25 [44.18, 50.54] & 34.65 [32.05, 37.32]
                & 93.41 [92.24, 94.59] & 89.76 [88.31, 91.23] \\
\bottomrule
\end{tabular}
\caption{Macro-F1 (\%) with 95\% bootstrap confidence intervals
(1{,}000 resamples).}
\label{tab:ci}
\end{table*}

\begin{table*}[t]
\centering
\small
\setlength{\tabcolsep}{4pt}
\renewcommand{\arraystretch}{1.0}
\begin{tabular}{lcccccccccc}
\toprule
\textbf{Model} & \textbf{CR} & \textbf{EE} & \textbf{NE} & \textbf{OP}
& \textbf{OG} & \textbf{OS} & \textbf{PE} & \textbf{TP}
& \textbf{NoE} & \textbf{OT} \\
\midrule
\multicolumn{11}{l}{\textit{Open-weight}} \\[2pt]
Deepseek-v4-flash & .82/.75 & .44/.39 & .80/.54 & .64/.56 & .39/.03 & .40/.03 & .30/.08 & .47/.12 & .95/.94 & .00/.00 \\
Deepseek-v4-pro   & .82/.72 & .51/.36 & .84/.67 & .70/.62 & .45/.25 & .44/.09 & .26/.16 & .56/.08 & .92/.89 & .00/.00 \\
Qwen-3.5 \textit{\small(397B)}        & .85/.76 & .53/.41 & .85/.73 & \textbf{.75}/.60 & .46/.22 & .52/.15 & \textbf{.43}/.07 & .63/.47 & .97/.85 & .06/.00 \\
Qwen-3.6-35b      & .85/.77 & .49/.40 & .76/.64 & .65/.57 & .28/.12 & .32/.06 & .30/.16 & .37/.00 & .94/.96 & .00/\textbf{.07} \\
Qwen3-14b         & .79/.69 & .39/.29 & .73/.67 & .65/.48 & .38/.10 & .24/.00 & .33/.06 & .26/.15 & .94/.95 & .00/\textbf{.07} \\
\midrule
\multicolumn{11}{l}{\textit{Closed-source}} \\[2pt]
GPT-6.1-sol       & .82/--  & \textbf{.55}/-- & .86/-- & .73/-- & \textbf{.53}/-- & .35/-- & .28/-- & \textbf{.82}/-- & \textbf{.99}/-- & .00/-- \\
Claude-Sonnet-5.5 & .78/.77 & .47/.48 & .79/.76 & .62/.60 & .30/.33 & .17/.23 & .19/.23 & .45/.55 & .97/.97 & .00/.00 \\
Qwen-3.6-Plus     & \textbf{.86}/.80 & .55/.49 & \textbf{.87}/.77 & .67/.62 & .45/.18 & \textbf{.51}/.03 & .40/.14 & .61/.31 & .96/.90 & .00/.00 \\
\bottomrule
\end{tabular}
\caption{Per-category F1 under \textbf{thinking} (left of slash) and \textbf{no-thinking} (right of slash). GPT-6.1-sol doesn't support no-thinking results. CR = Correct, EE = Entity Error, NE = Numeric Error, OP = Opposite, OG = Over-Generalization, OS = Over-Specific, PE = Partial Evidence, TP = Tortured Phrase, NoE = No Evidence, OT = Other. Best values in \textbf{bold}.}
\label{tab:per-category}
\end{table*}

\section{Use of AI Assistants}
This research was conducted with the assistance of AI tools for text refinement and coding support. 
% All scientific content and conclusionsare the responsibility of the authors.

\end{document}